\documentclass[letterpaper, 10 pt, conference]{ieeeconf}  

\IEEEoverridecommandlockouts                              
                                                 
\usepackage{graphics} 
\usepackage{epsfig} 
\usepackage{mathptmx} 
\usepackage{times} 
\usepackage{amsmath} 
\usepackage{amssymb}  
\usepackage{url}
\usepackage{romannum}
\usepackage{xcolor}
\usepackage{float}
\usepackage{url}
\usepackage[ruled,vlined]{algorithm2e}
\usepackage{array}
\usepackage{placeins}
\usepackage{adjustbox}
\usepackage{color}

\title{\LARGE \bf

Visual Place Recognition with Low-Resolution Images 

}

\author{Mihnea-Alexandru Tomiță$^{1}$, Bruno Ferrarini$^{1}$, Michael Milford$^{2}$, Klaus McDonald-Maier$^{1}$, Shoaib Ehsan$^{1,3}$
\thanks{$^{1}$Authors are with the School of Computer Science and Electronic Engineering, University of Essex, CO4 3SQ, United Kingdom
        {\tt\small matomi@essex.ac.uk, bferra@essex.ac.uk, kdm@essex.ac.uk, sehsan@essex.ac.uk}. }
\thanks{$^{2}$Michael Milford is with the School of Electrical Engineering and Computer Science, Queensland University of Technology, Brisbane, QLD 4000, Australia
        {\tt\small michael.milford@qut.edu.au}}
\thanks{$^{3}$Shoaib Ehsan is also with the School of Electronics and Computer Science, University of Southampton, SO17 1BJ, United Kingdom
        {\tt\small s.ehsan@soton.ac.uk}}
\thanks{This work is supported by the UK Engineering and Physical Sciences Research Council through grants EP/R02572X/1 and EP/P017487/1.}
}

\begin{document}

     \maketitle
    \thispagestyle{empty}
    \pagestyle{empty}

     \begin{abstract}

     Images incorporate a wealth of information from a robot's surroundings. With the widespread availability of compact cameras, visual information has become increasingly popular for addressing the localisation problem, which is then termed as Visual Place Recognition (VPR). While many applications use high-resolution cameras and high-end systems to achieve optimal place-matching performance, low-end commercial systems face limitations due to resource constraints and relatively low-resolution and low-quality cameras. In this paper, we analyse the effects of image resolution on the accuracy and robustness of well-established handcrafted VPR pipelines.  
     Handcrafted designs have low computational demands and can adapt to flexible image resolutions, making them a suitable approach to scale to any image source and to operate under resource limitations. This paper aims to help academic researchers and companies in the hardware and software industry co-design VPR solutions and expand the use of VPR algorithms in commercial products.
     
    \end{abstract}

      
    \section{Introduction}\label{introduction}
     With the advances in technology made in the last decade, image and video capturing devices became exceptional in reproducing a higher quality representation of our surroundings. Visual Place Recognition (VPR) utilises the visual information gathered from the camera to perform the localisation process. To achieve high place matching performance, VPR applications usually employ high-end systems and advanced cameras \cite{VPRsurvey}. However, low-end commercial products are computationally limited and have low-resolution cameras. Thus, the deployment of robust but computationally demanding VPR methods is restricted on such platforms, as identified in \cite{zaffar2019state, 8792942}. Hence, handcrafted VPR techniques are suitable to be deployed on resource constrained platforms, due to their computationally efficient nature. In addition to their low computational requirements, handcrafted VPR techniques can adapt to various image resolutions, which makes them attractive for VPR applications on resource-constrained platforms, with low-resolution cameras. Moreover, as a lower-resolution image is visually different from its high-resolution version  (refer to Fig. \ref{image_resolution}), this paper analyses the optimal image resolution for different handcrafted descriptors. Thus, the aim of this paper is to reduce the image resolution to facilitate VPR applications on resource-constrained commercial platforms. In summary, our contributions are as follows: 

    \begin{figure}[t]
            \centering
            \begin{tabular}{ c }

                \includegraphics[width=230pt]{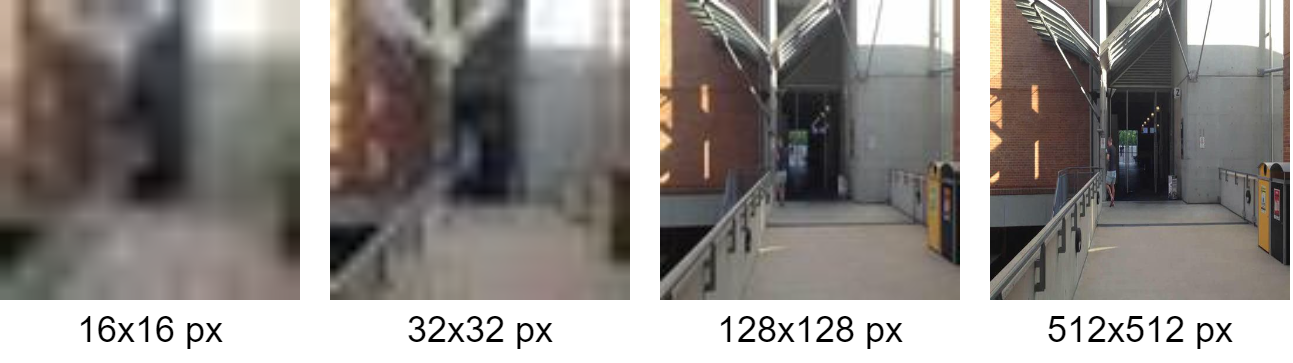}
                
            \end{tabular}
            \caption{The same image resized to various resolutions.}
            \label{image_resolution}
            \end{figure}
     
     \begin{itemize}
         \item An assessment of the performance of several well-established handcrafted VPR techniques on various image resolutions. We employ several datasets to enable a VPR performance comparison in real-world scenarios, under illumination, viewpoint and seasonal variation. 
         \item We report the total time required to perform VPR for each descriptor, showing how a reduced image resolution results in a more efficient VPR process. We also perform a trade-off analysis between performance and computation, showing the best descriptor that should be selected depending on the image resolution. 
     \end{itemize}

     The remainder of this paper is organised as follows: Section \ref{literature_review} presents the literature review. 
     Section \ref{experimental_study} presents the experimental setup, where the VPR time is discussed. We also present the performance metric employed, together with the selection of VPR techniques and datasets utilised in this study. Section \ref{results} presents the detailed results and analysis. The conclusions are presented in section \ref{conclusion}.

\begin{figure*}[t]
            \centering
            \begin{tabular}{ c c c c }

                \includegraphics[width=112pt, trim=8 8 8 8, clip]{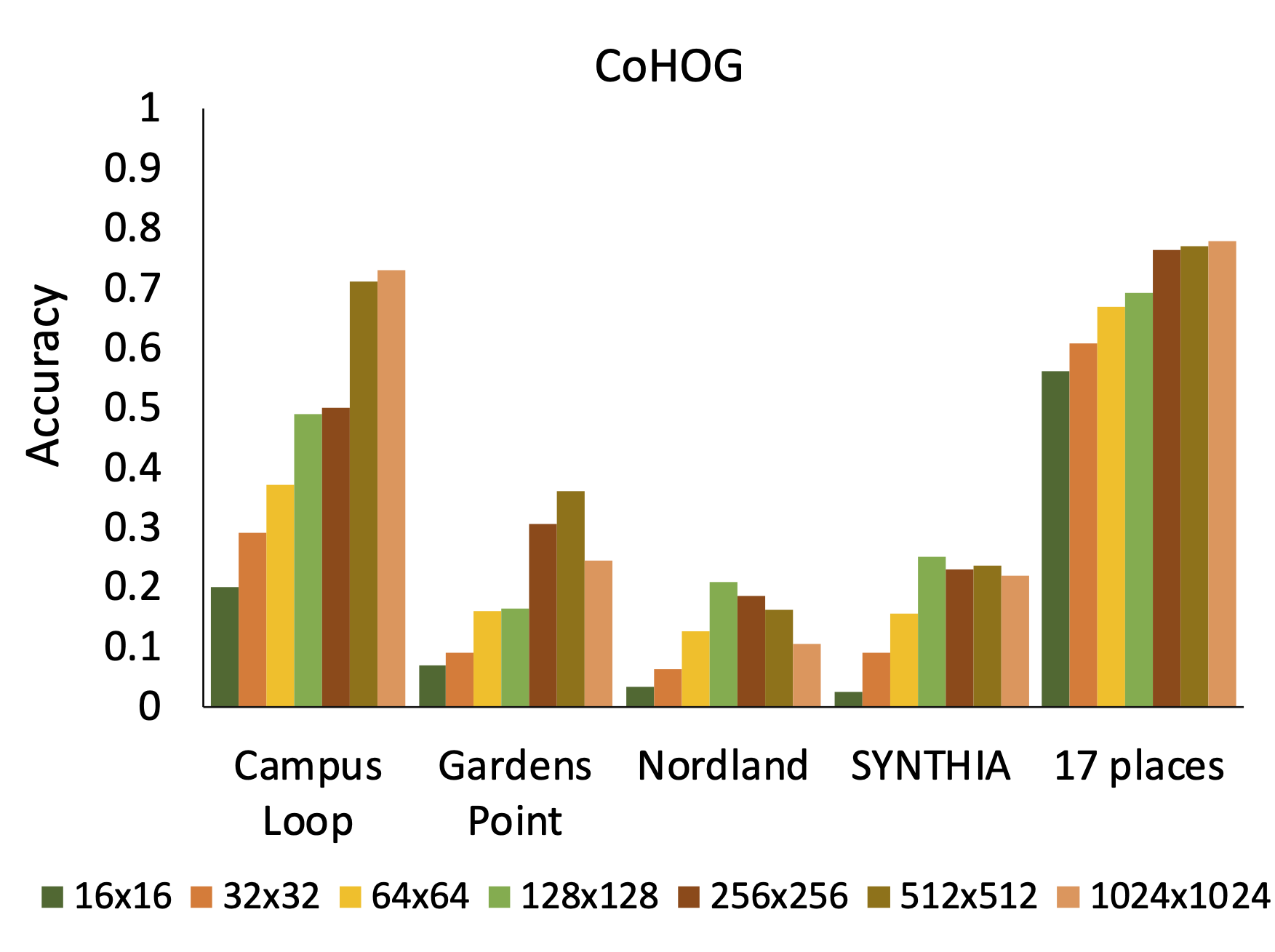} & 
                \includegraphics[width=112pt, trim=8 8 8 8, clip]{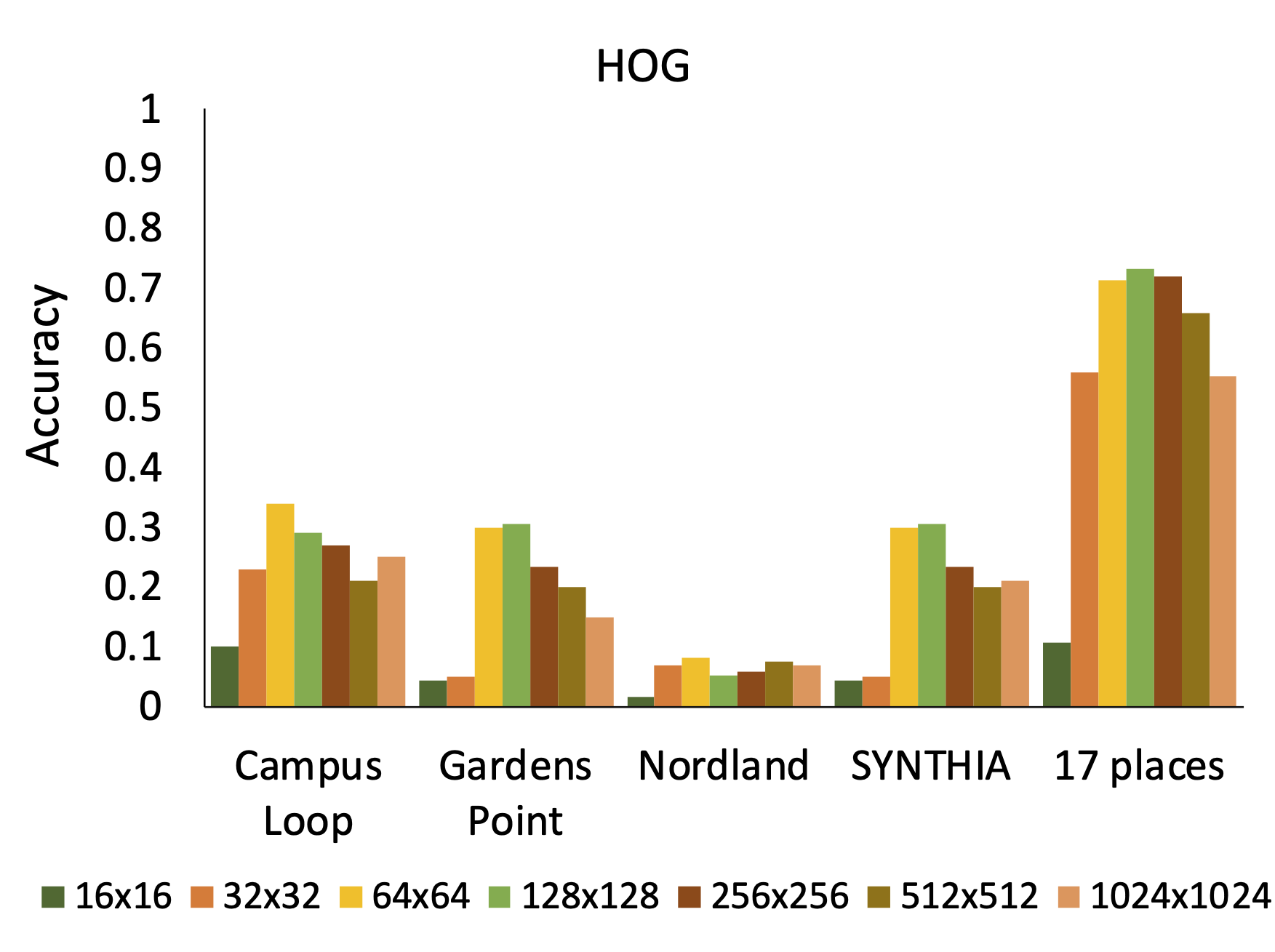} &
                \includegraphics[width=112pt, trim=8 8 8 8, clip]{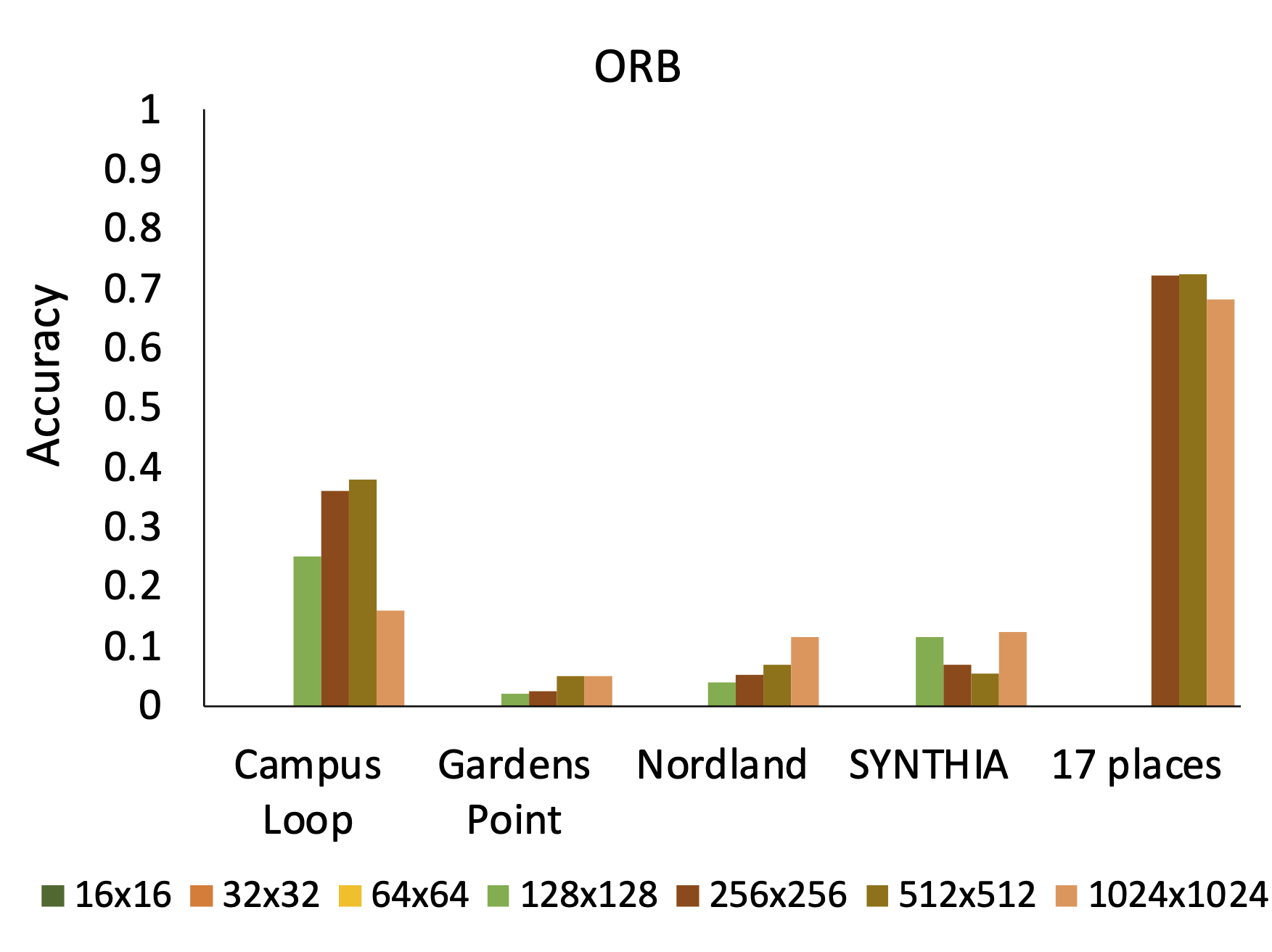}  &
                \includegraphics[width=112pt, trim=8 8 8 8, clip]{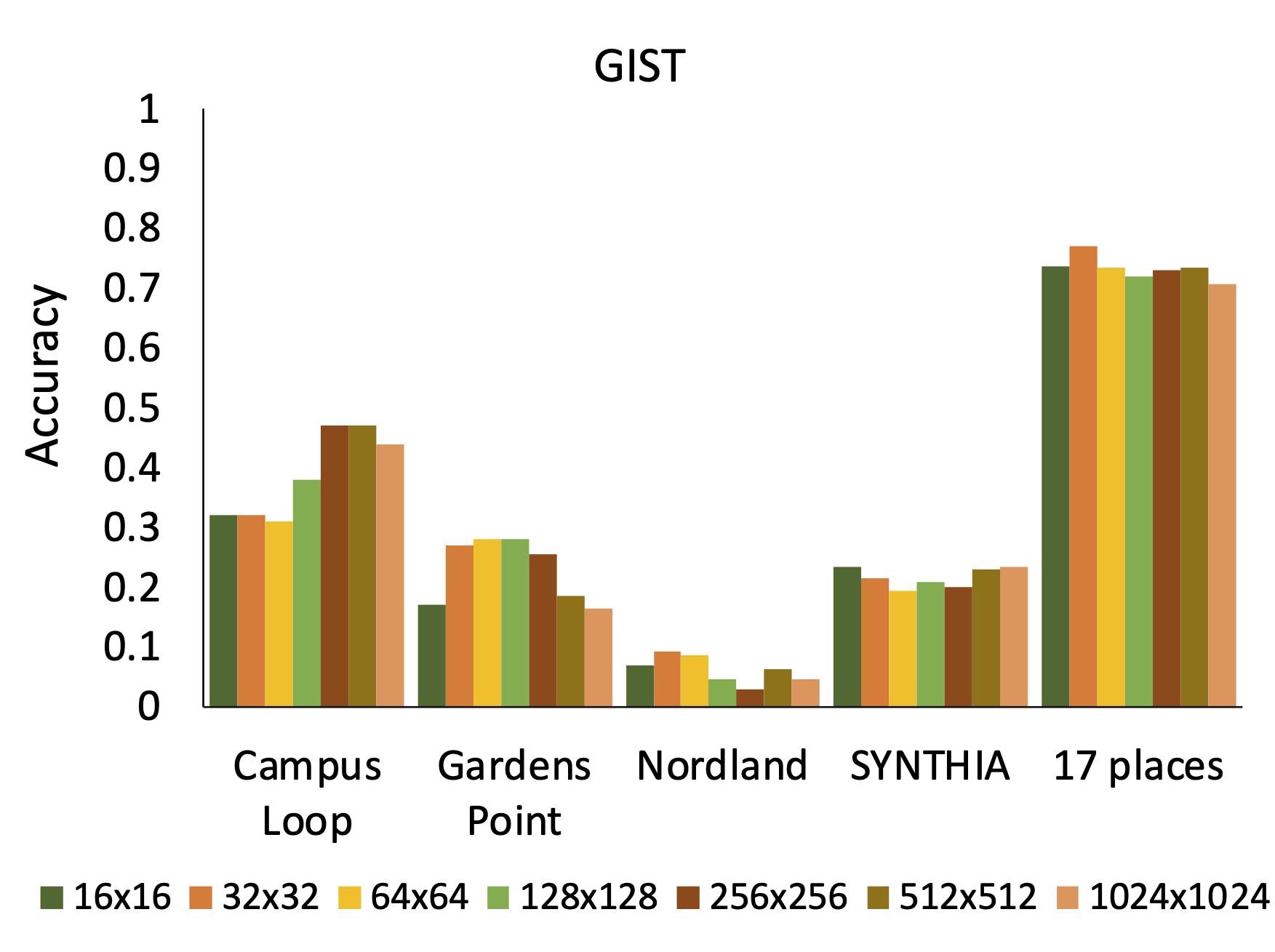}
                
            \end{tabular}
            \caption{The accuracy of all VPR techniques on each resized dataset.}
            \label{accuracy}
            \end{figure*}
     
     \begin{table}[t]
            \caption{The size of each dataset in Megabytes (MB) resized to various resolutions.} 
            \begin{adjustbox}{width=\columnwidth,center}
              \begin{tabular}{ |c|c|c|c|c|c|c|c| } 
            
            \hline
            & \multicolumn{7}{c|}{\textbf{Image Resolution [px]}} \\\cline{2-8}
            \textbf{Dataset} & \textbf{16x16} & \textbf{32x32} & \textbf{64x64} & \textbf{128x128} & \textbf{256x256} & \textbf{512x512} & \textbf{1024x1024}\\
            \hline
            17 places & 0.671 & 0.872 & 1.5 & 3.3 & 8.4 & 21.8 & 57.9\\
            \hline
            Campus Loop & 0.151 & 0.194 & 0.339 & 0.845 & 2.7 & 9.3 & 28.7\\
            \hline
            Gardens Point & 0.442 & 0.573 & 1 & 2.5 & 7.5 & 23.1 & 63.9\\
            \hline
            Nordland & 0.25 & 0.311 & 0.512 & 1.2 & 3.4 & 10.1 & 29.8\\
            \hline
             SYNTHIA & 0.296 & 0.379 & 0.647 & 1.5 & 4.6 & 16.2 & 56.9\\
            \hline
            \end{tabular}
            \end{adjustbox}

            \label{table:image_size}
            \end{table}

            \begin{figure}[t]
            \centering
            \begin{tabular}{ c }

                \includegraphics[width=185pt, trim=8 8 8 8, clip]{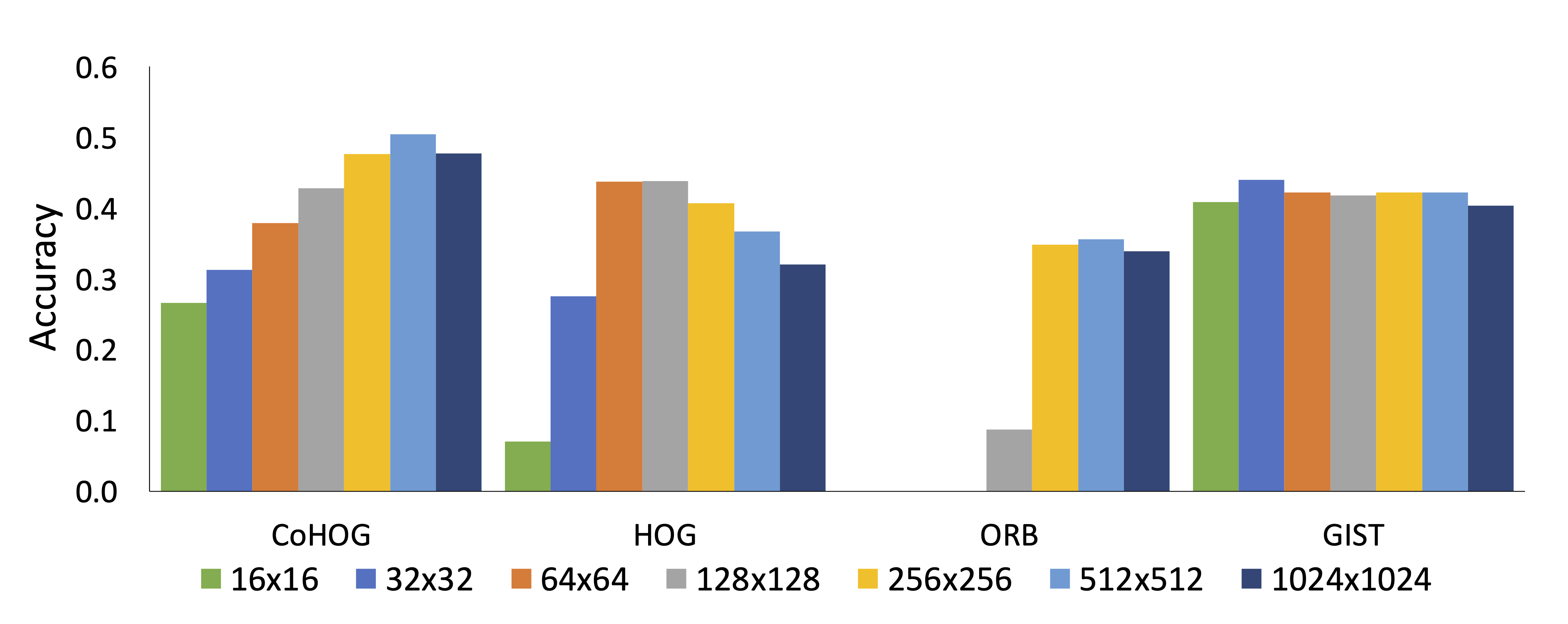}
                
            \end{tabular}
            \caption{The average accuracy of each technique on the combined datasets.}
            \label{average_accuracy}
            \end{figure}

            \begin{figure}[t]
            \centering
            \begin{tabular}{ c }

                \includegraphics[width=185pt]{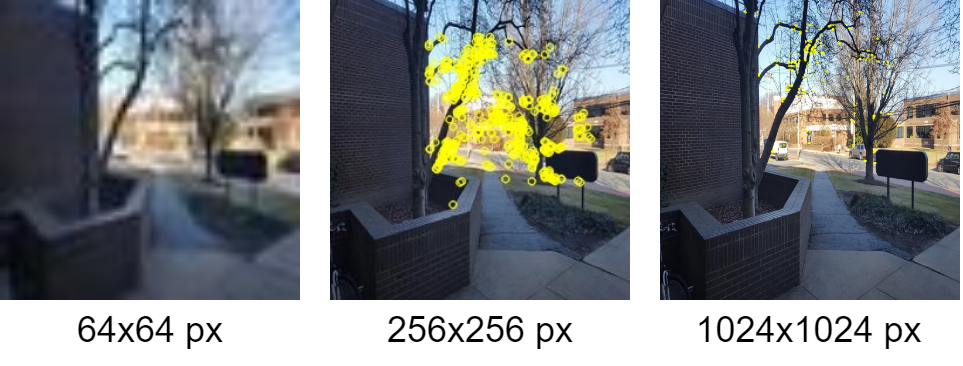}
                
            \end{tabular}
            \caption{Keypoints found in the same image at several distinct resolutions, as determined by ORB descriptor.}
            \label{ORB_keypoints}
            \end{figure}


                
    
    \section{Literature Review}   \label{literature_review}
      Prior to the deployment of deep-learning for VPR applications, handcrafted local feature descriptors were primarily utilised to solve VPR challenges. However, these cannot handle severe illumination changes in the environment. In contrast with local feature descriptors that analyse keypoints in images, global feature descriptors analyse the entire content of the image \cite{VPRsurvey}.


      A popular global feature descriptor is GIST \cite{oliva2001modeling} utilised in VPR applications such as \cite{murillo2009experiments, singh2010visual}. HOG \cite{dalal2005histograms} is a computationally efficient global descriptor, tolerant to appearance changes \cite{zaffar2021vpr}. In \cite{zaffar2020cohog}, the authors propose CoHOG, a compute-efficient, and training-free VPR system based on HOG descriptors, having good tolerance to lateral shifts.

      Local feature descriptors such as SURF \cite{SURF} and SIFT \cite{lowe2004distinctive,SIFT} have been widely used in VPR applications such as \cite{se2002mobile,andreasson2004topological}. 
      BRIEF \cite{calonder2011brief} has similar VPR performance with SIFT and SURF, albeit at a reduced encoding time. Bag-of-Words model (BoW) \cite{4270197} and Vector of Locally Aggregated Descriptors (VLAD) \cite{jegou2010aggregating} build an image descriptor of fixed length by aggregating local feature descriptors around centroids. BoW and VLAD can be used for VPR as shown in \cite{jegou2010aggregating} and \cite{8792942}, respectively. The authors of ORB \cite{rublee2011orb} propose a computationally-efficient descriptor, capable of performing real-time VPR. 

      In this work, the focus is mainly towards global feature descriptors, as local features descriptors are unable to detect keypoints in small images, as later shown in section \ref{perfromance}. Thus, they are not suitable to operate on small resolution images.


                

            \begin{figure*}[t]
            \centering
            \begin{tabular}{ c c c c }

                \includegraphics[width=112pt, trim=8 8 8 8, clip]{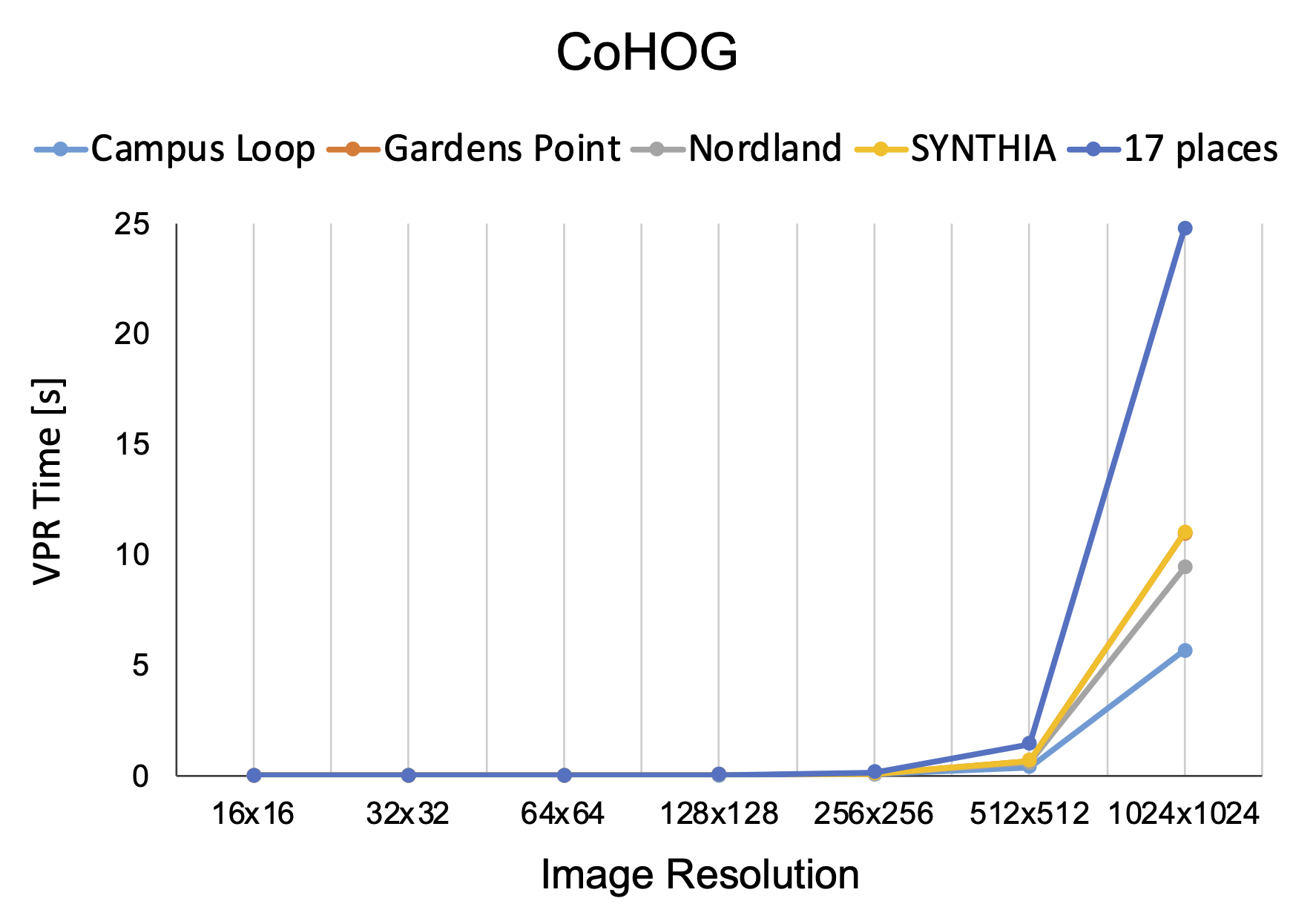} & 
                \includegraphics[width=112pt, trim=8 8 8 8, clip]{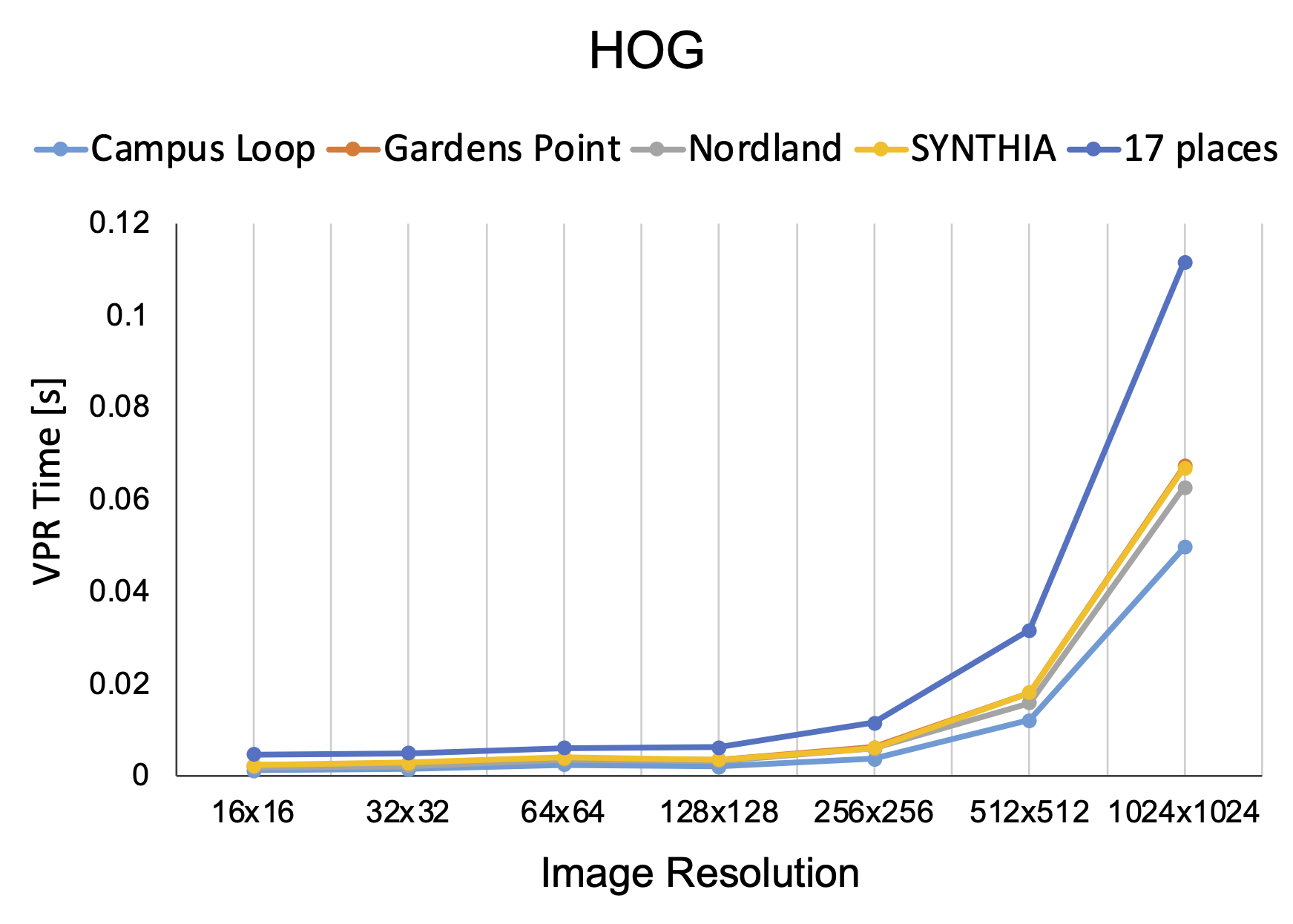} &
                \includegraphics[width=112pt, trim=8 8 8 8, clip]{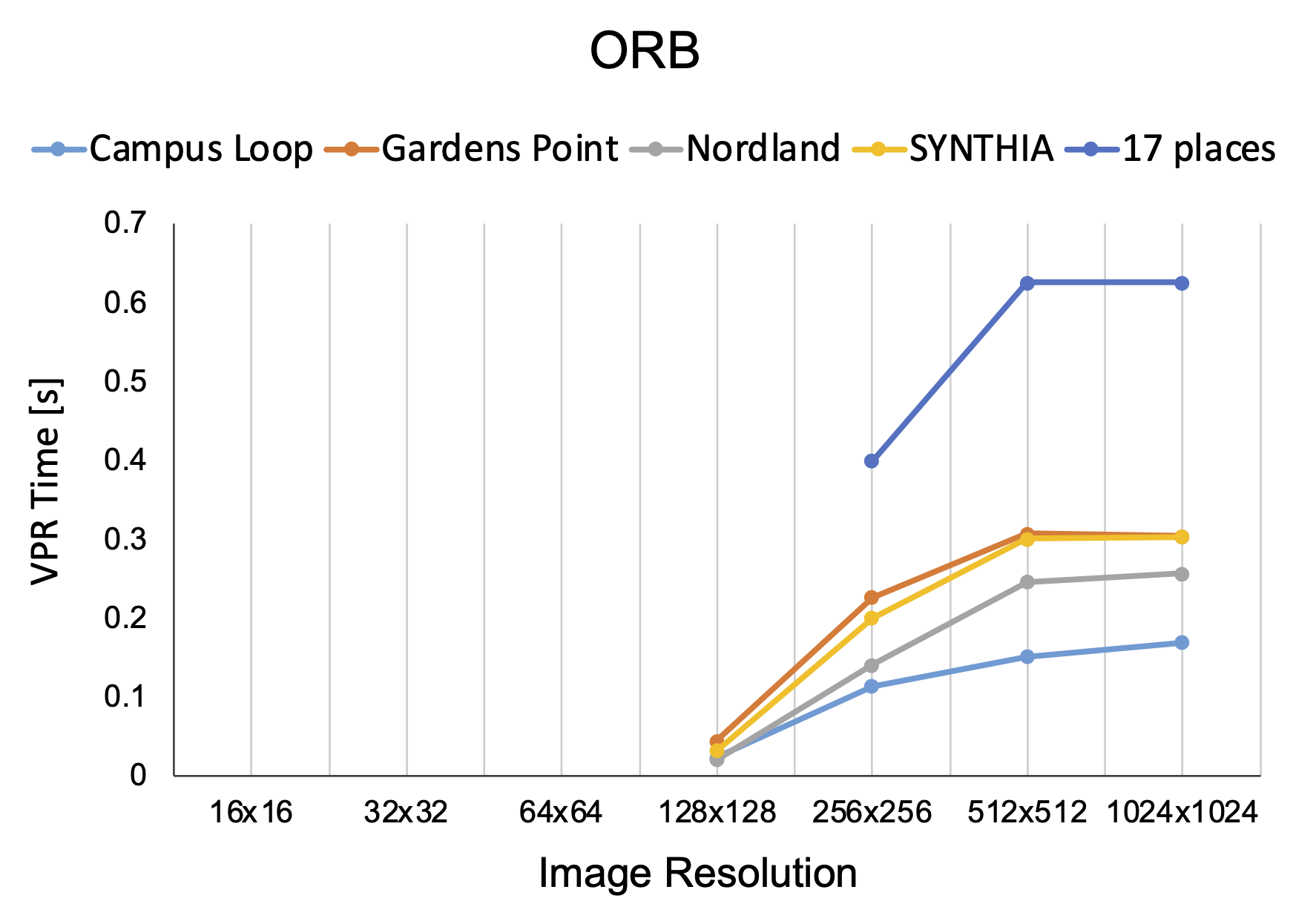}  &
                \includegraphics[width=112pt, trim=8 8 8 8, clip]{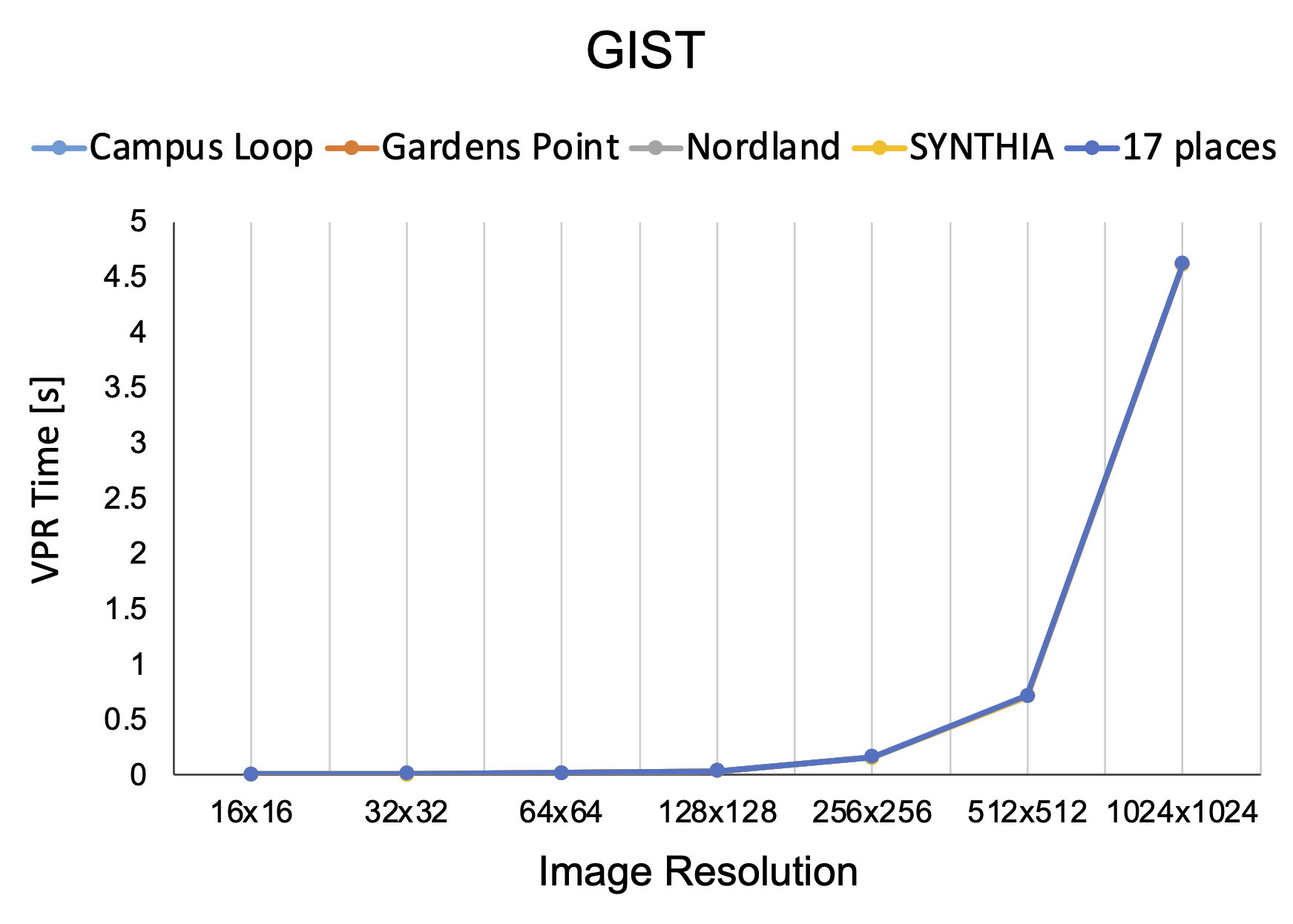}
                
            \end{tabular}
            \caption{The VPR time (refer to equation (\ref{eq:vprtime})) in seconds (s) of all VPR techniques on various image resolutions.}
            \label{VPR_time}
            \end{figure*}

\begin{figure*}[t]
            \centering
            \begin{tabular}{ c c c c }

                \includegraphics[width=112pt, trim=8 8 8 8, clip]{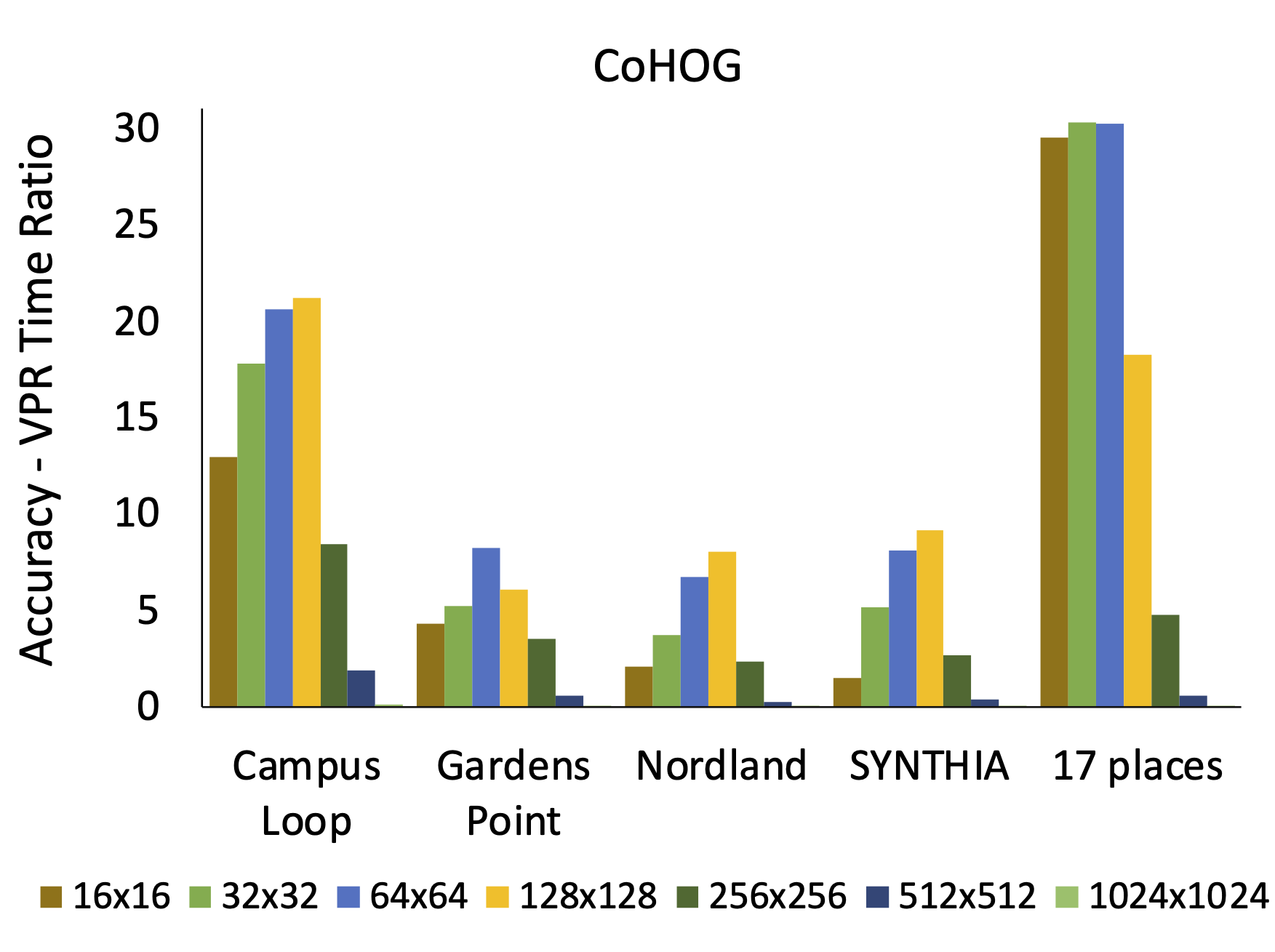} & 
                \includegraphics[width=112pt, trim=8 8 8 8, clip]{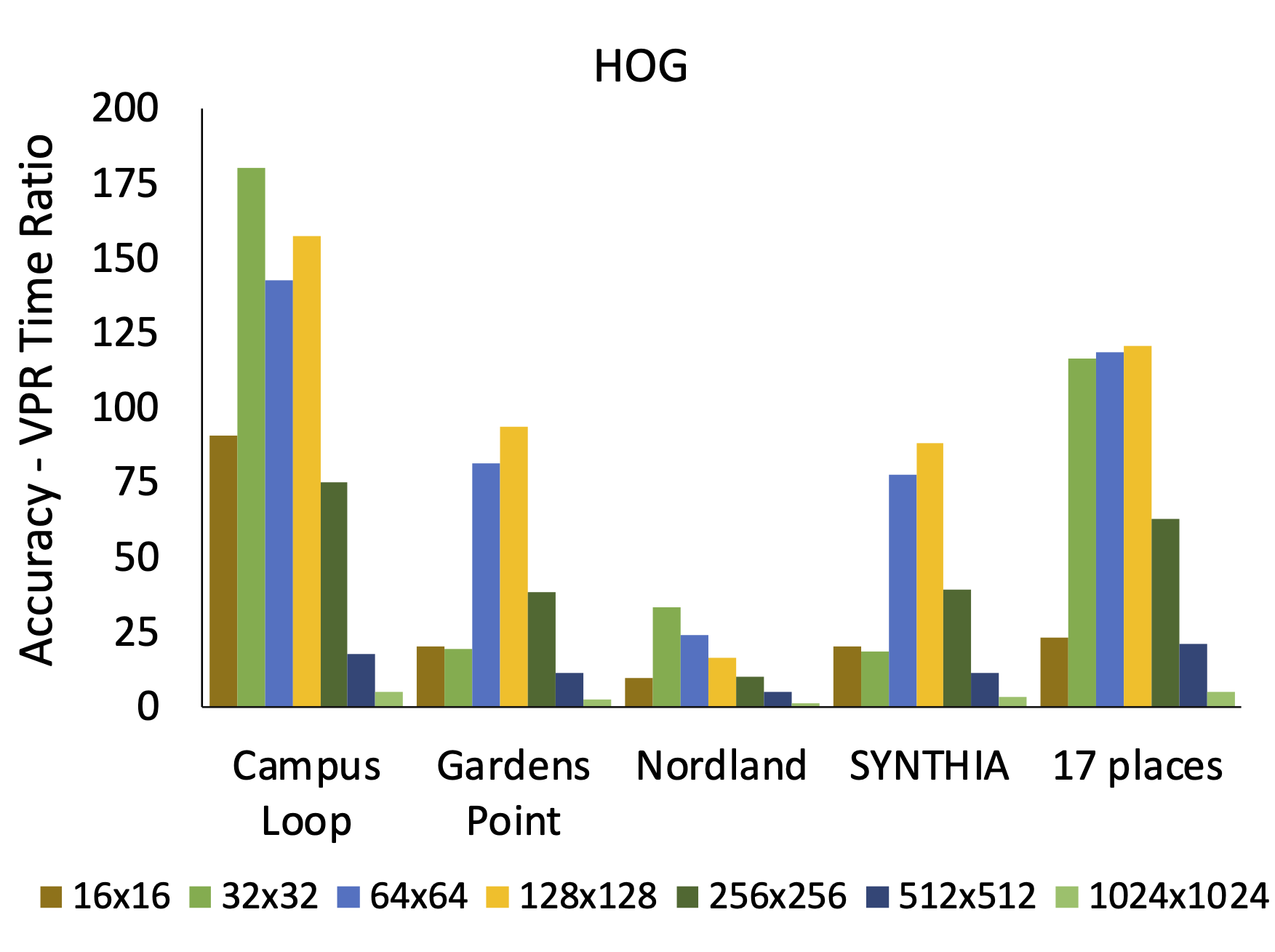} &
                \includegraphics[width=112pt, trim=8 8 8 8, clip]{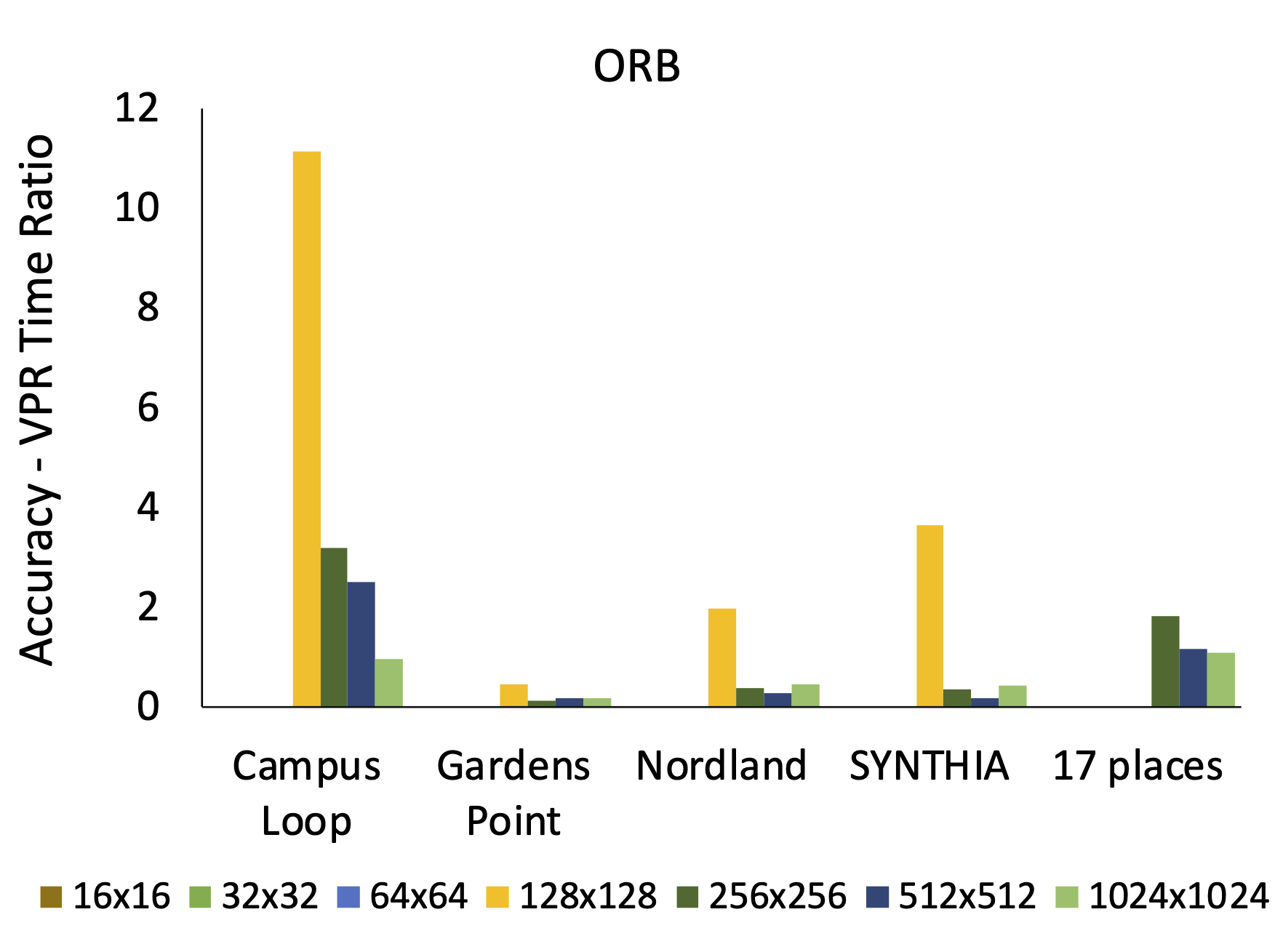}  &
                \includegraphics[width=112pt, trim=8 8 8 8, clip]{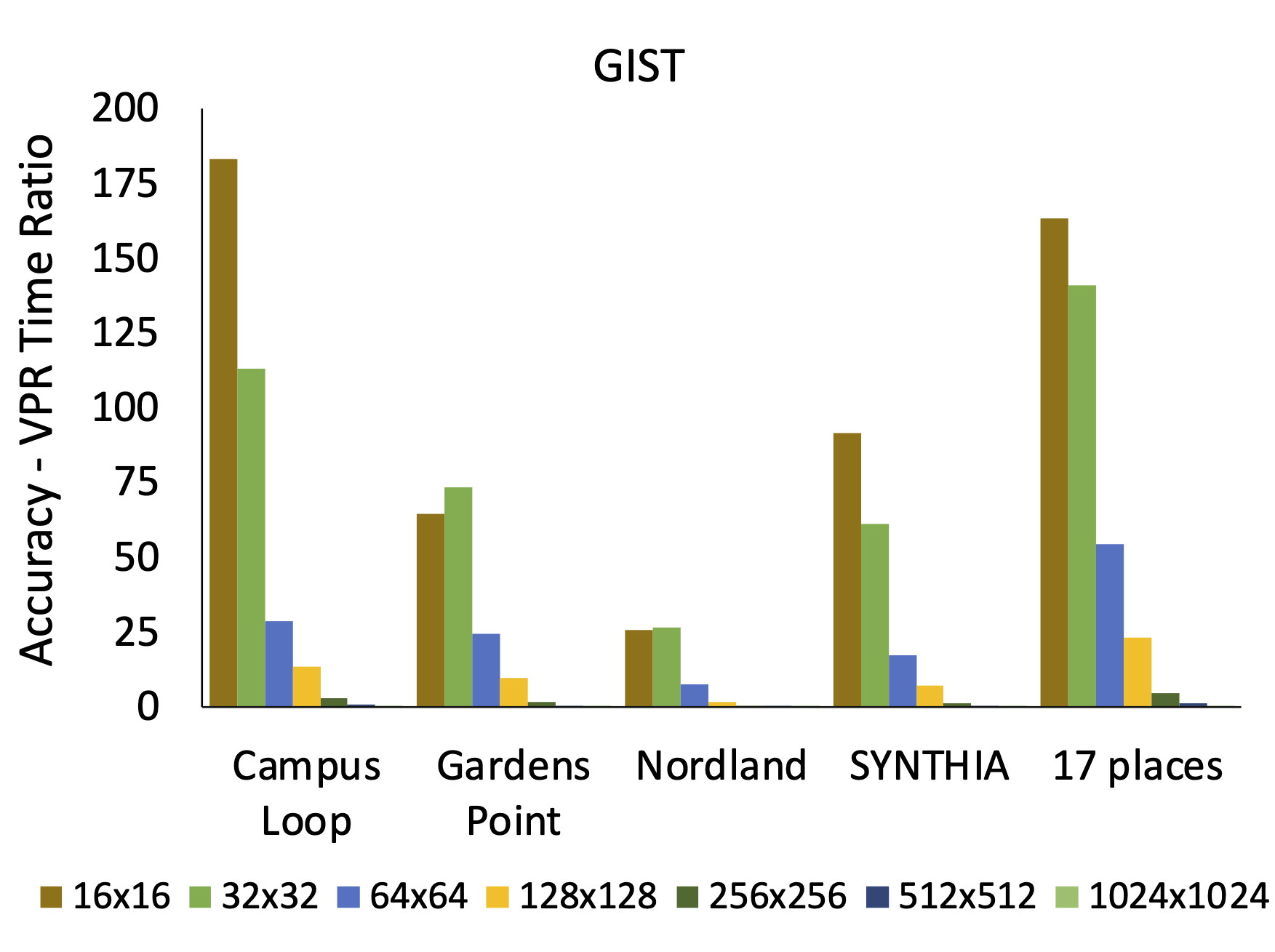}
                
            \end{tabular}
            \caption{The ratio between accuracy and VPR time of each technique on all resized dataset.}
            \label{ratio}
            \end{figure*}

  \begin{table}
            \caption{The encoding time in milliseconds (ms).} 
            \begin{adjustbox}{width=\columnwidth,center}
            \begin{tabular}{ |c|c|c|c|c|c|c|c| } 
            
            \hline
           \textbf{VPR} & \multicolumn{7}{c|}{\textbf{Image Resolution [px]}} \\\cline{2-8}
            \textbf{Technique} & \textbf{16x16} & \textbf{32x32} & \textbf{64x64} & \textbf{128x128} & \textbf{256x256} & \textbf{512x512} & \textbf{1024x1024}\\
            \hline
            CoHOG & 14.5 & 15.3 & 16.6 & 19 & 30.6 & 77.3 & 260.1 \\
            \hline
            HOG & 0.104 & 0.236 & 1.307 & 0.514 & 1.585 & 6.578 & 31.32 \\
            \hline
            ORB & - & - & - & 0.86  & 2.49 & 6.171 & 17.5 \\
            \hline
            GIST & 0.967 & 2 & 9.807 & 27.561 & 153.99 & 708.02 & 4618.1 \\
            \hline
            \end{tabular}
            \end{adjustbox}
            \label{table:encoding_time}
\end{table}
            
\begin{table}
            \caption{The average matching time in milliseconds (ms).} 
            \begin{adjustbox}{width=\columnwidth,center}
            \begin{tabular}{ |c|c|c|c|c|c|c|c| } 
            
            \hline
            \textbf{VPR} & \multicolumn{7}{c|}{\textbf{Image Resolution [px]}} \\\cline{2-8}
            \textbf{Technique} & \textbf{16x16} & \textbf{32x32} & \textbf{64x64} & \textbf{128x128} & \textbf{256x256} & \textbf{512x512} & \textbf{1024x1024}\\
            \hline
            CoHOG & 2.806 & 2.985 & 3.634 & 11.9 & 81.25 & 852.12 & 15398.19\\
            \hline
            HOG & 2.887 & 3.036 & 3.166 & 3.749 & 6.318 & 15.79 & 50.864 \\
            \hline
            ORB & - & - & - & 30.16  & 263.12 & 400.7 & 392.4 \\
            \hline
            GIST & 2.32 & 2.247 & 2.375 & 2.309 & 2.389 & 2.418 & 2.67 \\
            \hline
            
            \end{tabular}
            \end{adjustbox}
            \label{table:matching_time}
\end{table}

    \section{Experimental Setup} \label{experimental_study}
    \subsection{VPR Time}
      For low-end commercial products which are computationally limited, it is important to determine the optimal technique. Hence, in this paper we utilise the total time required to perform VPR ($t_{VPR}$) as a measurement of computational efficiency. The $t_{VPR}$ of each technique is determined by adding the encoding time $t_e$ with the matching time $t_m$ as follows:

        
        \begin{equation}\label{eq:vprtime}
            t_{VPR} = t_e + t_m,
            \end{equation} where $t_e$ refers to the amount of time that a VPR technique requires to compute the feature descriptor of an image and $t_m$ represents the time required to match the descriptor of a query image with all the reference descriptors in the map.
            
    \subsection{Performance Metric}
        To evaluate the VPR performance on various image resolutions, the percentage of correctly matched images is utilised, having the following formula:
\begin{equation}\label{eq:accuracy}
               Accuracy = \frac{N_c}{N_q}\;\text{,} 
        \end{equation} where $N_c$ represents the number of correctly matched query images and $N_q$ is the total number of query images. The accuracy has values in range [0,1], higher values denoting better VPR performance. 
    
    \subsection{VPR Techniques}\label{vprtechniques}
    
     A selection of four well-established VPR techniques have been employed in this work including: HOG \cite{dalal2005histograms}, CoHOG \cite{zaffar2020cohog}, ORB \cite{rublee2011orb} and GIST \cite{oliva2001modeling}. For HOG, a cell and block size of 16x16 pixels was utilised, with a total number of histogram bins of 9 \cite{zaffar2021vpr}. The remaining VPR techniques have been utilised as presented by their authors, with no additional changes being made to neither technique.
     Cosine similarity was used to match places for HOG and GIST. CoHOG was used with the built-in matching algorithm. Pair-wise local feature matching was employed for ORB.
     
    \subsection{Test Datasets}
    In this paper, five well-established VPR datasets are employed to present our findings. These are as follows: Campus Loop dataset \cite{merrill2018lightweight} consists of 100 query and 100 reference images, with a large amount of frames that contain viewpoint and seasonal variations; Gardens Point dataset \cite{sunderhauf2015performance} consists of 200 query (\textit{day\_left}) and 200 reference (\textit{night\_right}) images, with a focus on illumination and viewpoint variation; Nordland dataset \cite{nordland2013dataset} captures the drastic changes between seasons. In this paper, we have utilised 172 query images (summer) and 172 reference images (winter) of the Nordland dataset. SYNTHIA dataset \cite{SYNTHIA} is a simulated city-like environment that consists of 200 query and 200 reference images, taken in various weather, seasonal and illumination conditions. 17 places \cite{sahdev2016indoor} is an indoor dataset, whose images contain illumination and viewpoint variation. For this study, three locations have been selected entitled Arena, AshRoom and Corridor. Hence, this dataset consists of 457 query (\textit{day\_vme1}) and 434 reference images (\textit{night\_vme1}). All the datasets employed in the experiments have at least one corresponding reference image in the map to enable the use of the accuracy metric (refer to equation (\ref{eq:accuracy})).
    

    To enable a place matching performance comparison of each technique employed, the above mentioned datasets have been resized to several image resolutions (values presented in pixels (px)), ranging from 16x16 px to 1024x1024 px. Fig. \ref{image_resolution} presents some sample images taken from the Gardens Point \textit{day\_left} dataset resized to various image resolutions. Table \ref{table:image_size} presents the size in Megabytes of each resized dataset.

    \section{Results and Analysis}\label{results}

        \subsection{Place Matching Performance}\label{perfromance}
        The performance of all VPR techniques on every resized dataset is presented in Fig. \ref{accuracy}. In contrast with the VPR accuracy of HOG and GIST which peaks towards smaller images, CoHOG benefits from an increased image resolution. Moreover, as CoHOG is designed to handle lateral shifts in camera movement, this technique achieves high accuracy on 17 places and Campus Loop datasets, while utilising a higher image resolution than the rest of the techniques (1024x1024 px). This trend is also emphasized in Fig. \ref{average_accuracy}, which presents the average performance for each technique on all presented datasets, where the accuracy for each image resolution is weighted with regards to the number of images in the dataset. CoHOG achieves the highest place matching performance on datasets resized to 512x512 px.  For GIST, the highest accuracy is reported on the datasets resized to 32x32 px. HOG achieves similar levels of performance on both 64x64 px and 128x128 px resized datasets, as seen in Fig. \ref{average_accuracy}. It is important to note that ORB cannot work with small image resolutions, as previously mentioned in section \ref{literature_review}. This happens because no keypoints are detected in images, or the image is smaller than the descriptor patch. In our experiments, ORB cannot work with image resolutions of less than 128x128 px. Moreover, for 17 places dataset, ORB does not find any keypoints in image resolutions of less than 256x256 px. Fig. \ref{ORB_keypoints} shows the keypoint locations of ORB at different image resolutions. It can be seen that by reducing the image resolution to 64x64 px, ORB fails to detect any of the previously identified keypoints in the presented environment.
        


        \subsection{Analysis on the Time Required to Perform VPR}\label{VPR_Time_Analysis}
        This sub-section performs an analysis on the total time required to perform VPR. Table \ref{table:encoding_time} presents the encoding time $t_e$ and Table \ref{table:matching_time} presents the matching time $t_m$ of all VPR techniques. Moreover, the VPR time (refer to equation (\ref{eq:vprtime})) of every technique is presented in Fig. \ref{VPR_time}. We have previously discussed in sub-section \ref{perfromance} that CoHOG achieves increased levels of place matching performance utilising a higher image resolution. However, as the matching time $t_m$ of CoHOG is really high when utilising a higher image resolution, its VPR time is drastically increased, as seen in Fig. \ref{VPR_time} on the 17 places dataset. When utilising an image resolution of 128x128 px and above, GIST achieves high encoding times when compared to the remaining VPR techniques, as reported in Table \ref{table:encoding_time}. 
        %
        %
        Also, GIST has a short and insensitive to resolution matching time (Table \ref{table:matching_time}). GIST's descriptor does not change with the image size as opposed to the others. In particular, CoHOG and ORB detect more local regions as the resolution increases, yielding a longer matching time.
        %
        %
        Moreover, the $t_{VPR}$ of GIST does not differ significantly from one dataset to another, as shown in Fig. \ref{VPR_time}. 
        %
        Overall, GIST should be selected for VPR applications with a focus on fast processing times. However, if the aim is towards VPR performance, CoHOG should be utilised instead.

        \subsection{Performance and Computation Trade-off Analysis}
         As utilising a lower image resolution generally results in a decrease in $t_{VPR}$ (refer to Fig. \ref{VPR_time}), this section performs a trade-off analysis between VPR performance and time. The ratio between the accuracy and VPR time of each technique on all resized datasets is presented in Fig. \ref{ratio}. As previously mentioned in sub-section \ref{perfromance}, CoHOG generally achieves higher place matching performance while using larger image resolutions, albeit at a considerable increase in $t_{VPR}$. Thus, in comparison with VPR techniques such as HOG and GIST which perform better whilst utilising a lower image resolution, the ratio for CoHOG is considerably lower. HOG achieves the highest ratio using either 32x32 px or 128x128 px, depending on the dataset. For GIST, the highest ratio is obtained on the 16x16 px resized datasets, with the exception of Gardens Point and Nordland, where this is achieved on the 32x32 px resized datasets, as seen in Fig. \ref{ratio}.




        


                
            
    \section{Conclusions} \label{conclusion}

   This paper investigates the effects of image resolution on the place matching performance of several well-established handcrafted VPR techniques using several well-established VPR benchmark datasets. The results presented show that VPR performance tends to degrade as resolution decreases but, at the same time, improves the speed of place recognition. The results also show that different VPR approaches are impacted differently by resolution. In particular, ORB - a local feature extractor - is unusable on low-resolution images. An extension of this work will include more realistic datasets where the existence of a reference image is not guaranteed, whilst also providing an in-depth investigation of the benefits of low-resolution images for visual privacy. Thus, VPR could be performed using low-resolution images in environments where delicate visual information is present, such as faces in crowded environments and car plate numbers.

   
   


    {
    \small
    \bibliographystyle{ieeetr}
    \bibliography{root}
    }
    
\end{document}